\documentclass[letterpaper]{article} 
\usepackage{aaai23}  
\usepackage{times}  
\usepackage{helvet}  
\usepackage{courier}  
\usepackage[hyphens]{url}  
\usepackage{graphicx} 
\usepackage{natbib}  
\usepackage{caption} 
\usepackage{algorithm}
\usepackage{algorithmic}

\usepackage{newfloat}
\usepackage{listings}
\DeclareCaptionStyle{ruled}{labelfont=normalfont,labelsep=colon,strut=off} 
\floatstyle{ruled}
\newfloat{listing}{tb}{lst}{}
\floatname{listing}{Listing}
\usepackage{booktabs}
\title{Different Affordances on Facebook and SMS Text Messaging Do Not Impede Generalization of Language-Based Predictive Models}
\author { 
Tingting Liu\textsuperscript{\rm \equalcontrib,1,2},
Salvatore Giorgi\textsuperscript{\rm \equalcontrib,1,2},
Xiangyu Tao\textsuperscript{\rm 3},
Sharath Chandra Guntuku\textsuperscript{\rm 2}, \\
Douglas Bellew\textsuperscript{\rm 1},
Brenda Curtis\thanks{Share the senior authorship.}\textsuperscript{\rm ,1,}\thanks{Corresponding author.},
Lyle Ungar $^\dagger$\textsuperscript{\rm ,2}\\
}
 \affiliations {
     \textsuperscript{\rm 1} National Institute on Drug Abuse \\
     \textsuperscript{\rm 2} University of Pennsylvania \\
     \textsuperscript{\rm 3} Fordham University \\
     \{tingting.liu, sal.giorgi, doug.bellew, brenda.curtis\}@nih.gov, xtao16@fordham.edu,  \{sharathg, ungar\}@cis.upenn.edu
}

\RequirePackage{subfiles}

\begin{document}

\maketitle

\begin{abstract}
Adaptive mobile device-based health interventions often use machine learning models trained on non-mobile device data, such as social media text, due to the difficulty and high expense of collecting large text message (SMS) data. Therefore, understanding the differences and generalization of models between these platforms is crucial for proper deployment. We examined the psycho-linguistic differences between Facebook and text messages, and their impact on out-of-domain model performance, using a sample of 120 users who shared both. We found that users use Facebook for sharing experiences (e.g., leisure) and SMS for task-oriented and conversational purposes (e.g., plan confirmations), reflecting the differences in the affordances. To examine the downstream effects of these differences, we used pre-trained Facebook-based language models to estimate age, gender, depression, life satisfaction, and stress on both Facebook and SMS. We found no significant differences in correlations between the estimates and self-reports across 6 of 8 models. These results suggest using pre-trained Facebook language models to achieve better accuracy with just-in-time interventions. 
\end{abstract}

\section{Introduction}

Language reflects users' psychology and can be used to understand and predict mental health conditions (i.e., \citealp{de2013predicting}). While language from social media such as Facebook has been widely used~\cite{eichstaedt2018facebook,jaidka2018facebook, liu2022linguistic},
text messaging (Short Message Service or SMS) is emerging as a new platform for detecting mental health conditions and delivering interventions (e.g., depression:~\citealp{liu2021relationship}, 
loneliness: ~\citealp{liu2022head}). This also opens possibilities for Just-in-Time Adaptive Interventions (JITAIs) to deliver physical and mental health support based on an individual's changing state and environment~\cite{nahum2018just}.

Most JITAIs are designed for smartphones, but current NLP models are primarily trained on social media, not SMS. Acquiring large-scale SMS for fine-tuning models is difficult and expensive. Transferring pre-trained Facebook models to SMS is thus common \cite{liu2021relationship}. But the impact of cross-platform language differences is unclear. Our evaluation aims to scientifically quantify the differences and ensure the successful transfer of current NLP models trained on social media sites into potential JITAIs using SMS.

\begin{figure}[!tb]
    \centering
    \includegraphics[width=1\columnwidth]{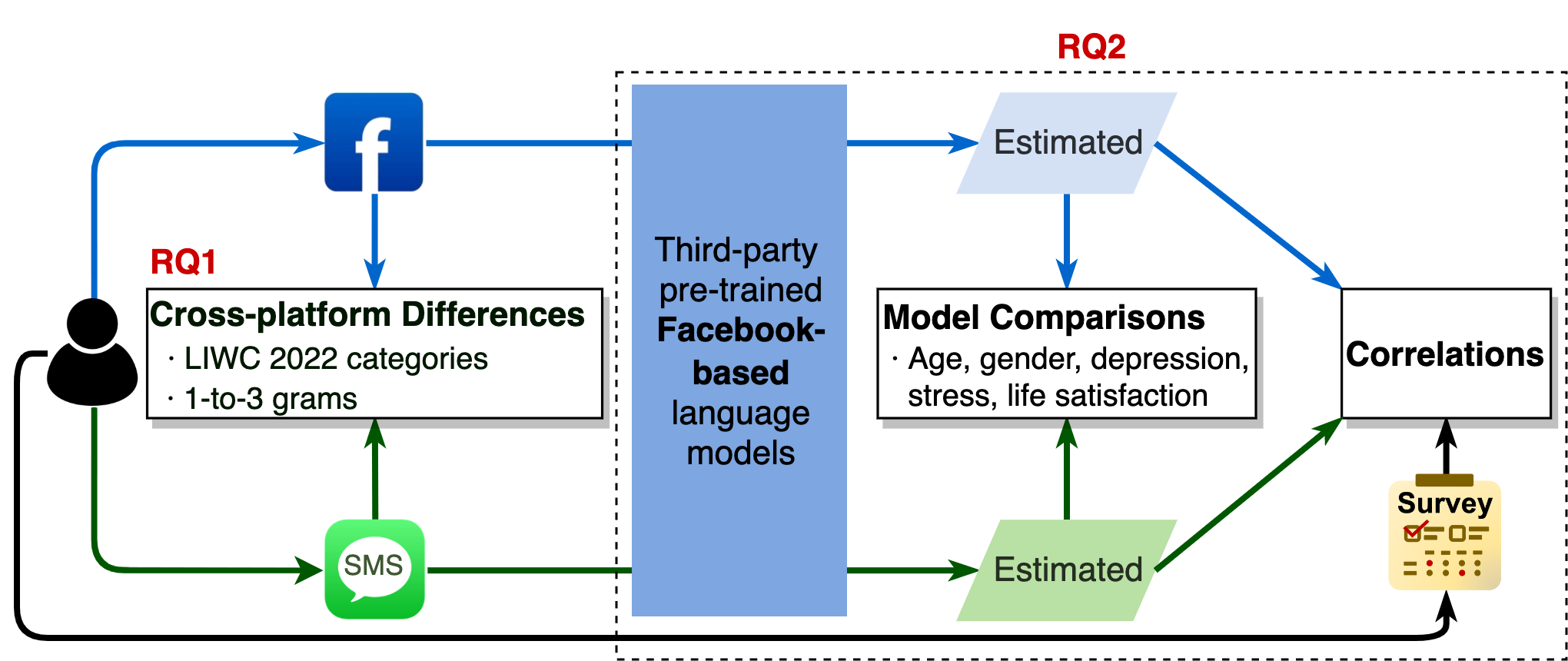}
    \caption{For the same users, we: (1) compared differences between Facebook and SMS language, and (2) evaluated the in- vs. out-of-domain efficacy of language models in predicting users' self-reported psychological traits.}
    \label{fig:flow}
\end{figure}
Our research aims to explore: 
\textbf{RQ1} the distinctions in language between Facebook and SMS, and \textbf{RQ2} to evaluate the efficacy of language models derived from Facebook data in predicting psychological traits when applied to SMS. To achieve this, we utilize \emph{the same cohort of users} who have provided their Facebook language, SMS, and psychometric self-reports (e.g., demographics, depression; Figure \ref{fig:flow}).

\paragraph{Contributions} Our contributions are: (1) showing clear Facebook vs. SMS distinctions in language use for the same users; (2) evaluating the two platforms by training/validating within and across domains; and (3) laying the foundation for NLP model transfer to SMS with within-user comparisons.

\section{Background}
Language use varies across contexts. How can the language use of the same person differ in SMS and Facebook? Although, to our knowledge, no study has compared these two, some research has compared Facebook status updates with direct messages (DM), a private message on Facebook resembles SMS. 
~\citet{doi:10.1177/0261927X12456384} observed that sharing positive emotions is associated with self-presentational concerns in Facebook status updates but not DM, noting the difference between communication on public and private channels.
~\citet{10.1111/jcom.12106} also identified various goals and motivations for self-disclosures in Facebook status updates and DMs. Status updates relate to more social validation, self-expression, and relief, while DMs relate to relationship development and social maintenance.

If individuals use Facebook and SMS for different functions, it is still unclear how well models trained on Facebook posts will perform when applied to SMS data. Preliminary work showed that linguistic model predictions change across platforms. For example, \citet{seabrook2018predicting} examined the association between depression and emotional word expressions on Facebook and Twitter and found different patterns. On Facebook, the instability of negative emotion words only predicts depression, whereas on Twitter, the variability of negative emotion words reflects the severity of depression. However, as 
most cross-platform comparisons are made across user groups; individual differences (e.g., demographic) may cause these variations more than language choices. As cross-platform generalizations are expected to lead to model performance degradation, only a few studies have conducted same-user cross-platform comparisons~\cite{jaidka2018facebook, guntuku2019understanding}, no previous research has quantified these differences across Facebook and SMS within the same users. Our paper aim to draw more attention to these differences and provide possibilities for actionable improvements to conduct more precise predictions for future JITAIs plans.

\section{Data}
\label{sec:data}

\paragraph{Participants} Participants were recruited online via Qualtrics as part of a larger national survey ~\cite{tao2023covid}. Each consenting participant (1) lived in the U.S., (2) was over 18 years old, (3) shared Facebook status updates, (4) installed the open-source mobile sensing application AWARE~\cite{ferreira2015aware} on their Android phones, (5) wrote at least 500 words across platforms (Facebook and SMS apps), and (6) completed a survey which contains questions on age, gender\footnote{We only analyze binary male/female gender,a limited and problematic sense of gender, due to limited data and the limitations of our gender estimation model. Three participants reporting a non-binary gender were excluded from the gender analysis.}, depression, life satisfaction, and stress. Our final sample included 120 participants ($M_{age}$ = 36.46, 69\% female).\footnote{\label{supplement}See Supplement for full details on participant recruitment, demographics, survey-based measures, and text-based estimates, at \url{https://github.com/TTRUCurtis/Facebook-vs-SMS-language}.} Table \ref{table stats} shows usage differences between Facebook and SMS.\footnote{\label{clean}Extensive cleaning was automatically applied (i.e., no human in the loop) to the keystroke data to remove any sensitive PII data. See Supplement for details. To fairly compare the Facebook data to the keystroke data, we applied the same cleaning pipeline to both.}

\begin{table}[!h]
\small
\centering
\renewcommand{\arraystretch}{1.25} %
\begin{tabular}{lcccccc}

\hline
 & \multicolumn{3}{c}{Words}  & \multicolumn{3}{c}{Posts} \\
\cmidrule(lr){2-4} \cmidrule(lr){5-7}
  & Med. & Mean & SD & Med. & Mean & SD    \\ 
\hline
FB & 12,800 & 26,652 & 37,924 & 1,279 & 2,193 & 2,599  \\  
SMS & 3,607 & 7,881 & 11,693  & 331 & 711 & 961    \\
\hline
\end{tabular}
\caption{Posts and word count statistics per platform (Med. = median and SD = standard deviation).}
\label{table stats}
\end{table}


\paragraph{Survey-Based Measures} For each participant, we collected self-reported age, gender, depression, stress, and life satisfaction via surveys used as gold-standard measures. We measured depression via the Patient Health Questionnaire (PHQ-9; ~\citealt{kroenke2001phq}), life satisfaction via Cantril's Ladder~\cite{cantril1965pattern}, and stress via Cohen's Perceived Stress Scale~\cite{cohen1983global}.\footref{supplement}

\paragraph{Text-Based Estimates} We employed off-the-shelf text-based models to estimate age, gender~\cite{sap2014developing}, depression~\cite{schwartz2017dlatk}, stress~\cite{guntuku2019understanding}, and life satisfaction~\cite{jaidka2020estimating}. All models were developed in previous studies and trained on Facebook status updates to predict survey-based self-reports via lexical features (i.e., bag-of-words or bag-of-topics models).\footref{supplement} 

For this study, we also trained RoBERTa-based models~\cite{liu2019roberta} on the data sets used in the original papers. These models were trained for depression, life satisfaction, and stress only, as we cannot access the original data used to train the age and gender models. Since this paper is not aimed to build state-of-the-art classifiers, we used the same model pipeline across depression, life satisfaction, and stress: (1) we extracted user-level RoBERTa embeddings using the penultimate layer, (2) reduced the dimensions of the resulting 768 dimension embedding (using non-negative matrix factorization) to 128 dimensions~\cite{v-ganesan-etal-2021-empirical}, and (3) applied a $\ell_2$ regularized Ridge regression with $\alpha=1$ (chosen via nested cross-validation). The RoBERTa-based models had similar accuracy to the lexical models.\footref{supplement} 
\section{Methods}
\label{sec:methods}

\paragraph{RQ1: Cross-platform Differences} We first tokenized the Facebook status updates and SMS data, using a tokenizer designed for social media data~\cite{schwartz2017dlatk}. We considered both 1-to-3 grams and the Linguistic Inquiry and Word Count (LIWC) 2022 dictionary~\cite{boyd2022development}. LIWC has been widely used in psychological sciences (e.g., ~\citealp{eichstaedt2018facebook}) and LIWC 2022 consists of 102 manually curated categories by psychologists. From both Facebook and SMS data, we extracted 1-to-3 grams and created a binary outcome variable for each participant to indicate which platform they were on. We then calculated effect size using Cohen's \emph{d} values between platforms and conducted a logistic regression using n-grams to predict the binary platform indicator in order to calculate statistical significance (\emph{p} values). 
Next, we extracted all LIWC 22 categories from each user's Facebook and SMS data. To calculate differences, we computed paired sample $t$-tests for each LIWC category between Facebook and SMS. All significance thresholds were adjusted using a Benjamini-Hochberg False Discovery Rate (FDR) correction~\cite{benjamini1995controlling}. 

\begin{table}[t]
\centering
\resizebox{\linewidth}{!}{%
\begin{tabular}{llc} 
\toprule
\multicolumn{3}{c}{Facebook}    \\ 
Category & Top frequent words & \emph{t} \\ 
\hline
 Leisure &	\emph{fun, weekend, play} &	14.65 \\
 Determiners	& \emph{the, a, my, this} &	9.63  \\
 Quantities & \emph{all, day, some, more} &	7.91 \\
 Power&	\emph{own, order, power, president} &	7.77\\ 
 Emotion	& \emph{love, good, happy, :), fun}	&7.35 \\
\bottomrule
\multicolumn{3}{c}{SMS}   \\ 
Category & Top frequent words & \emph{t} \\ 
\hline
 Auxiliary verbs &	\emph{is, have, be, was}&	-20.26  \\
 Communication &	\emph{thank, say, thanks, said, tell}	&-17.92 \\
 Discrepancy &	\emph{can, want, would} &	-14.90\\
 Assent&\emph{	yes, ok, yeah, okay}&	-14.82\\
 2nd person &	\emph{you, your, you're, u} &	-13.27  \\
\bottomrule
\end{tabular}
}
\caption{Paired \emph{t}-tests results of LIWC 2022 categories, showing top categories which differ between Facebook and SMS. All results are statistically significant at $p < 0.001$ after Benjamini-Hochberg FDR correction.}
\label{table LIWC}
\end{table}

\paragraph{RQ2: In vs. Out of Domain Estimates} Here we performed three tasks to answer this from two different approaches: First, \emph{Task 1} applied off-the-shelf models to both the Facebook and SMS data to evaluate in-and across-domain estimates and their generalization, 
and \emph{Task 2} examined which linguistic features were driving the differences in estimates in \emph{Task 1}. \emph{Task 3} opted not to use off-the-shelf models in \emph{Task 1} and \emph{2}. Instead, it involved training and assessing predictive models within and across each domain.

\emph{Task 1}: For each participant, we estimated age, gender, depression, life satisfaction, and stress from Facebook and SMS text using the text-based models described above. We then correlated the estimates with the gold-standard survey-based measures for both the lexical and embedding-based models. A statistical bootstrap test was used to assess differences in correlations between SMS-based estimates and Facebook-based estimates. 

\emph{Task 2}: To identify features driving lexical-based model estimates in both domains, we investigated feature importance $i$, which is defined as:
\begin{equation}
     i(f) = w_f\big(freq_{FB}(f) - freq_{SMS}(f)\big).
\label{eq feat import}
\end{equation}             
Here $w_f$ is the weight of the feature $f$ in the depression model, $freq_{*}(f)$ is the frequency of feature $f$ in either the Facebook (FB) or SMS domain. 

\emph{Task 3}: Finally, instead of using off-the-shelf models, we trained and evaluated predictive models within and across each data set. To do this, we trained models to predict our five outcomes (age, gender, depression, life satisfaction, and stress) using both text sources from the same person as training and testing data sets: (1) train on FB / test on FB, (2) train on FB / test on SMS, (3) train on SMS / test on SMS, and (4) train on SMS / test on FB. We used a leave-one-out cross-validation setup when training and testing within the same text domain (1 and 3). When testing across text domains (2 and 4), we trained a model using one text source, applied the model to the other text source (producing estimates of our 5 outcomes), and correlated those estimates with self-reports.

\section{Results}
\label{sec:results}

\begin{figure}[t]
    \centering
    \includegraphics[width=0.9\columnwidth]{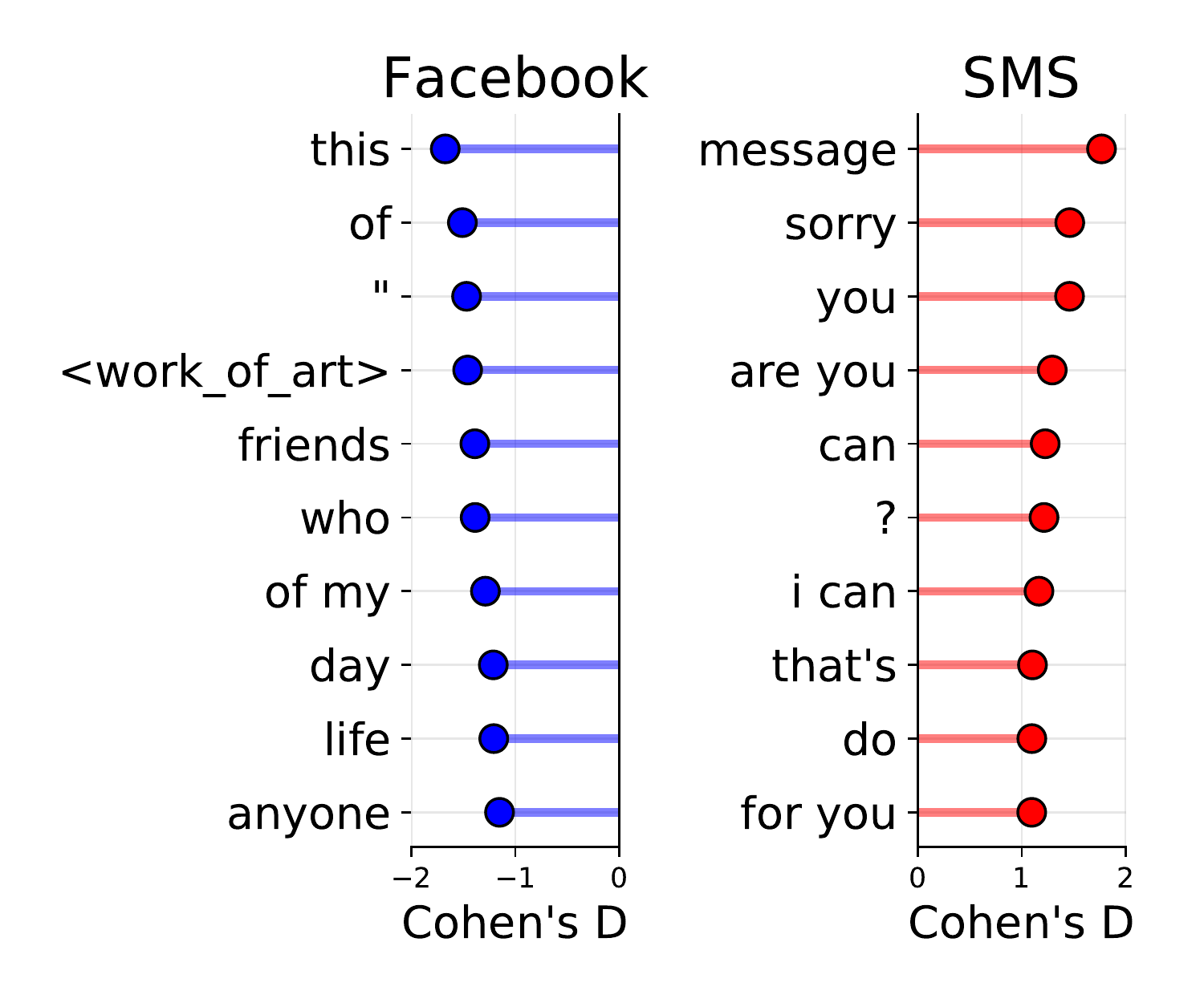}
    \caption{1-to-3 grams most correlated with Facebook vs. SMS, statistically significant at $p < 0.05$ after Benjamini-Hochberg FDR correction. Cohen's $d$ = effect size measuring Facebook vs. SMS differences. Angle brackets: spaCy annotated named entities (e.g., $<$work of art$>$).}
    \label{fig:ngram}
\end{figure}

\paragraph{RQ1: Cross-platform Differences}
As seen in Table~\ref{table LIWC} and Figure~\ref{fig:ngram}, people preferred to discuss leisure activities, share pleasant feelings (LIWC positive emotion) and contents (e.g., books, songs), and express their motivations (LIWC power) on Facebook. People used more conversational language (LIWC communication), more second-person pronouns, and were more task-oriented (e.g., actions, LIWC verbs, plan confirmations) in SMS. See Supplement\footref{supplement} for additional n-gram correlations that provided more context. 

\begin{table}[b]
\centering
\resizebox{.45\textwidth}{!}{
\begin{tabular}{lcccc}\toprule
 & \multicolumn{2}{c}{Lexical Models} & \multicolumn{2}{c}{Embedding Models} \\ \cmidrule(lr){2-3}\cmidrule(lr){4-5}
 & Facebook & SMS & Facebook & SMS \\ \hline
Age & .68 & .45$^*$ & - & - \\
Gender$^\dagger$ & .91 & .80$^*$ & - & - \\
Depression & .36 & .29 & .25 & .08 \\
Life Satis. & .21 & .14 & .31 & .31 \\
Stress & .21 & .18 & .21 & .23 \\\bottomrule
\end{tabular}
}
\caption{Pearson correlations (or $^\dagger$ accuracy) between language estimates and self-reports. $^*$ Significant difference in bootstrapping test between SMS and Facebook correlations.}
\label{tab:model estimated}
\end{table}

\begin{figure}[t]
    \centering
    \includegraphics[width=1\columnwidth]{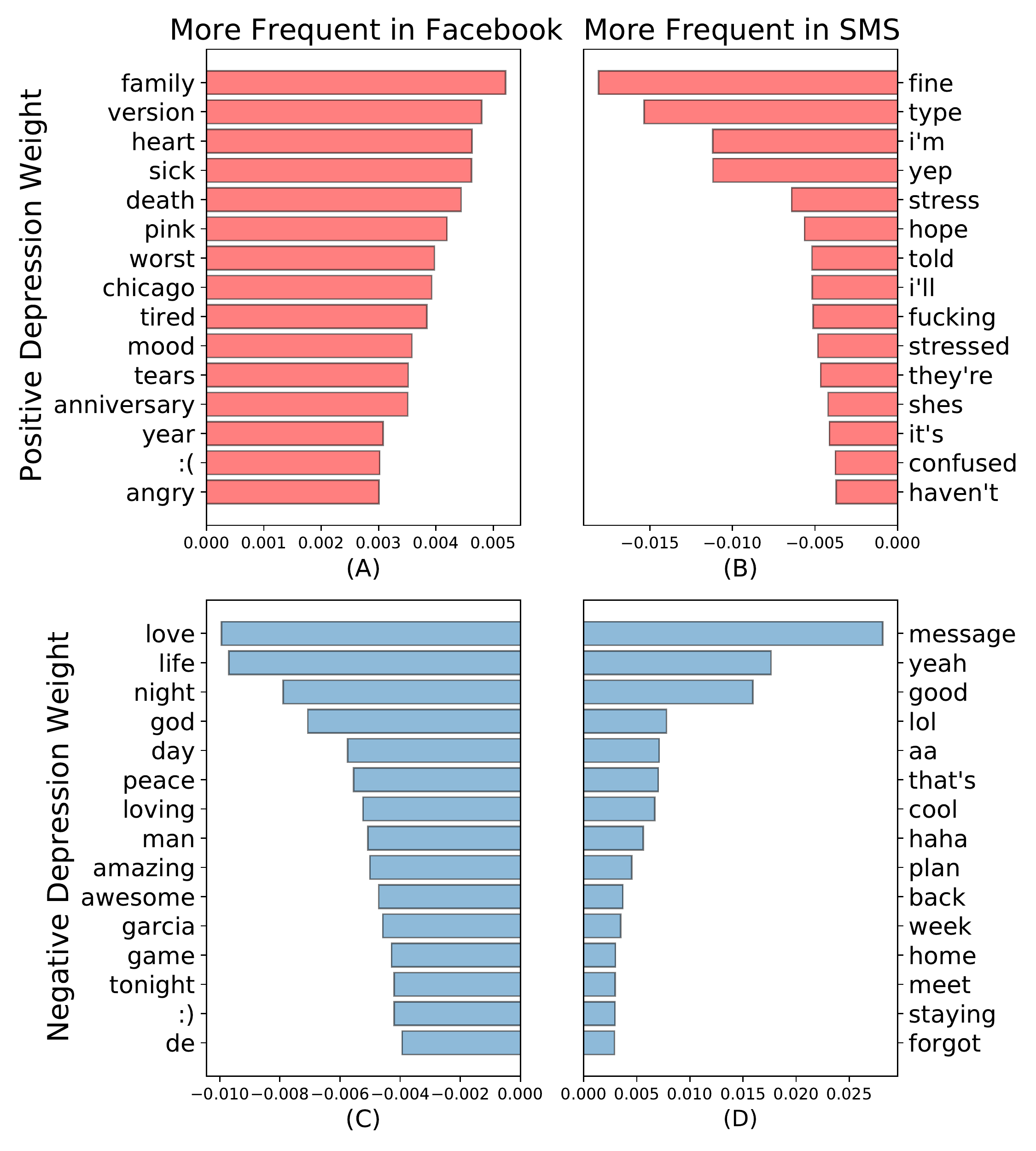}
    \caption{Feature importance results, as defined by the product of the depression model word weight and the difference in Facebook vs SMS word usage frequency. Top row (red bars; A and B) are positively weighted words in the depression model, while the bottom row (blue bars; C and D) are negatively weighted words. Left column (A and C) is more frequency words in Facebook (i.e., positive frequency difference), while the right column (B and D) contains words more frequent on SMS (i.e., negative frequency difference).}
    \label{fig:bar plots}
\end{figure}

\paragraph{RQ2: In vs. Out-of-Domain Estimates} In Table \ref{tab:model estimated}, we found that in-domain estimates from Facebook data predicted self-reports at rates similar to those in the original papers \footref{supplement} from which the models were built (\emph{Task 1}).
When predicting self-reports from SMS-based estimates (i.e., out-of-domain), we observed a drop in prediction accuracy across all lexical models and 1 out of 3 embedding models. However, the differences between the Facebook correlation with self-report and the SMS correlation with self-report were not statistically different (using a bootstrapping test) except for those for age and gender (where SMS does not perform as well as Facebook). In Figure \ref{fig:bar plots}, we further investigated the drop in performance by examining feature importance (\emph{Task 2}). Here we identified features reflecting language style, such as more use of contractions (``i'll'', ``i'm'', ``they're'', ``she's'', ``haven't''), driving the SMS depression estimates, and features about content, experience, and life events (``family'', ``sick'', ``anniversary'') driving the Facebook depression estimates.
In Table \ref{tab:within and across domains}, we presented the results from training and evaluating models within and across domains. Facebook-trained models have higher in-domain accuracy, and SMS-trained models have higher out-of-domain accuracy. Again, using a bootstrapping significance test, we did not see significant differences between the correlation of Facebook and self-reports versus SMS and self-reports (in both in- and cross-domain tasks). 



\begin{table}[!tb]
\centering
\resizebox{.45\textwidth}{!}{
\begin{tabular}{lcccc} \toprule
 & \multicolumn{2}{c}{In Domain} & \multicolumn{2}{c}{Cross Domain (train/test)} \\ \cmidrule(lr){2-3}\cmidrule(lr){4-5}
 & FB & SMS & FB/SMS & SMS/FB \\ \hline
Age & .61 & .52 & .40 & .50 \\
Gender$^\dagger$ & .75 & .74 & .63 & .73 \\
Depression & .25 & .09 & .15 & .32 \\
Life Satis. & .19 & .07 & .25 & .29 \\
Stress & .24 & .12 & .32 & .38 \\ \bottomrule
\end{tabular}
}
\caption{Within and across platform evaluation. Pearson correlations (or $^\dagger$ accuracy) between language estimates and self-reports. In Domain models are evaluated using leave-one-out cross validation. 
}
\label{tab:within and across domains}
\end{table}

\section{Conclusion}
\label{sec:conclusions}
Our study, based on data from \textit{the same users}, shows
: (1) individuals disclose different aspects of their lived experiences on Facebook and SMS
, (2) two platforms generate similar  mental health estimates, both within and across domains, whether using off-the-shelf models trained on Facebook data (Table \ref{tab:model estimated}) or models built specifically on the paper's dataset (Table \ref{tab:within and across domains}).
Consistent with past findings, Facebook usage reflects the need to belong and self-presentation~\cite{nadkarni2012people}, leading to more content sharing and opinion expression; whereas 
SMS is used for phatic communication to maintain social relationships and for informal discussions~\cite{fibaek2016s}, leading to more confirmations and conversational features.
Our data, derived from the same users, indicates that cross-platform differences can be attributed to language rather than demographics. Despite the linguistic differences, our findings suggest that predictions from both platforms are similar. 

\section*{Broader Impact}
Our findings have important implications. Firstly, our research highlights the variations in psycho-linguistic features between Facebook and SMS, thus warranting further investigation of downstream applications. Secondly, future researchers can build predictive models on large-scale social media language and apply them to SMS, which may offer a new approach to address the cost-accuracy trade-off in the context of just-in-time interventions on mobile devices.

This study involves human subjects and was approved by the Institutional Review Board (IRB). The data used in this study raise ethical concerns such as handling sensitive personal information (PII) and thus, we have taken measures to securely store, clean, and analyze the data, further data sharing is not possible\footref{clean}. We use social media, SMS data, and machine learning methods to estimate sensitive attributes like depression. Such estimates can have both positive and negative implications, ranging from providing support to causing discrimination. We must use them with caution. 

\section*{Acknowledgments}
This study was supported by the Intramural Research Program of the NIH, National Institute on Drug Abuse (ZIA-DA000632). The authors report no conflict of interest.

\bibliography{aaai23}

\maketitle

\section{Study Recruitment}

Participants were recruited online via the Qualtrics Panel as part of a larger national survey on mental health, substance use, and COVID-19. To qualify, consenting participants must be 18 years or older, U.S. residents, and Facebook users. To ensure they are active Facebook users, participants must have posted at least 500 words across their lifetime status updates and posted at least 5 posts within the past 180 days to be included in the study~\cite{eichstaedt2021closed}. 2,796 participants finished an initial survey, including questions about socio-demographics, and physical and mental health (e.g., depression, stress, and life satisfaction). This pool of participants has been used to study loneliness and alcohol use ~\cite{bragard2022loneliness} and COVID-related victimization ~\cite{tao2023covid}. The study received approval from the Institutional Review Board at our institution.

After completing this survey, participants were invited to install the open-source mobile sensing application AWARE~\cite{ferreira2015aware,nishiyama2020ios} on their mobile phones. This application collects mobile sensor information (e.g., movement, app usage, and keystroke data). A total of 300 participants installed and ran the AWARE app for 30 days. Keystroke data is only available for Android users (N = 192), thus we excluded 108 iPhone users. We only consider the Google, Verizon, and Samsung messaging apps as our keystroke text messaging sources, hereafter referred to as SMS data.  69 users who wrote less than 500 words within the 30-day study period were further excluded.
Finally, since the models used in this study are trained on monolingual English, 3 participants were removed due to mostly Spanish status updates. Thus, 120 participants entered our data analysis ($M (SD)_{age}$ = 36.46 (9.74), range: 18-65-year-old; 69\% female; the highest level of education: 57\% have four-year Bachelor’s degree or higher; household income: 49\% $> \$60,000$).

\section{Text Based Measures}

\paragraph{Age and Gender} We applied an age and gender predictive lexica~\cite{sap2014developing} built over a set of Facebook users who self-disclosed age and gender and shared their Facebook status updates. The final model predicted age with a Pearson \emph{r} = 0.86 and and binary gender with an accuracy = 0.90.

\paragraph{Depression} This model ~\cite{schwartz2017dlatk} was built on roughly 28,000 Facebook users who consented to share their Facebook data and answered the depression facet of neuroticism in the ``Big 5" personality inventory, a 100-item personality questionnaire (the International Personality Item Pool proxy to the
NEO-PI-R~\citep{goldberg1999broad}). This model resulted in a prediction accuracy (Pearson \emph{r}) of 0.386. Please see the original paper for full details. The RoBERTa-based model has a 10-fold cross-validation predictive accuracy of Pearson \emph{r} = 0.36.

\paragraph{Life Satisfaction} This model was built on roughly 2,700 Facebook users who consented to share their Facebook data and answered a life satisfaction questionnaire (see Cantril's Ladder below; ~\citealt{jaidka2020estimating}). The model was built using a set of 2,000 LDA topics and produced a prediction accuracy of Pearson \emph{r} = 0.26. The RoBERTa-based models have a 10-fold cross-validation predictive accuracy of Pearson \emph{r} = 0.29.

\paragraph{Stress} Similar to the depression and life satisfaction models, the stress model~\cite{guntuku2019understanding} was built over a set of Facebook users who answered Cohen's Perceived Stress inventory (see description below). Again, 2,000 LDA topics were used as features in a 10-fold cross-validation setup. The final model's accuracy was Pearson \emph{r} = 0.31. The RoBERTa-based models have a 10-fold cross-validation predictive accuracy of Pearson \emph{r} = 0.33.

\section{Survey Based Measures}

\paragraph{Depression}
Frequency of depression symptoms in the past two weeks are accessed via the 9-item Patient Health Questionnaire (PHQ-9; ~\citealt{kroenke2001phq}; e.g.,  ``Little interest or pleasure in doing things'') with response options ranged from 0 (Not at All) to 3 (Nearly Everyday). 

\paragraph{Life Satisfaction}
Life satisfaction is measured via Cantril's Ladder, which asks respondents to identify their current step on a ladder with steps numbered from zero at the bottom to 10 at the top, with the top representing the the best possible life, and the zero representing vice versa ~\cite{cantril1965pattern}.

\paragraph{Stress}
Stress is measured by Cohen’s Perceived Stress Scale (PSS; ~\citealt{cohen1983global}). A sample item is “In the last month, how often have you been upset because of something that happened unexpectedly?” Response options ranged from 0 (Never) to 4 (Very often).

\section{Keystroke Data and Text Cleaning}
The AWARE mobile sensing app logs each non-password keystroke on Android phones across all apps (e.g., text messages and search engine entries). These logs are stored one character at a time and include modifications such as deletions and auto-correct. For example, if a user searched ``Talyor Swift'' in a search engine, AWARE would log separate database entries for ``T'', ``Ta'', ``Tal'', etc. If the same user misspelled ``Talyor'' while typing, AWARE would also log the misspelling and the delete key; for example ``T'', ``Ta'', ``Tai'', ``Ta'' (i.e., a \emph{backspace} occurred), ``Tal'', etc. This presents a unique challenge when dealing with possibly sensitive information. 

While the main goal of cleaning Personal Identifiable Information (PII) is to enable non-trusted sources to access the collected data by removing PII, a secondary goal is to replace the PII with a tag indicating what kind of data has been removed to allow deeper analysis. Basic cleaning of each string was done in several stages. The first was to remove PII data that was structurally identified by the device itself as either a password field or a phone number. The second stage was to use \emph{spaCy's} Name Entity Recognizer (NER) and to replace all flagged entities with their category label. The third stage was to check against a list of common data formats using regular expressions using a modified version of CommonRegex\footnote{\url{https://github.com/madisonmay/CommonRegex}}. We noted that these category labels were ignored by our tokenizer and not used in the downstream analyses in the present study. 

Cleaning keystroke data which changes 1 character at a time; however, contains an extra challenge over standard complete string cleaning. Detection of partial PII data that doesn't yet match a known form (but will eventually) is required. We accomplished this by rolling future data back through the previous data in two stages. The first stage was that, each time when the completion of a new token at the end of a string was detected, we applied the replacement information, or lack thereof, back through the previous strings until the beginning of that token (there may be incomplete tokens which match NER that were not necessary to replace based on subsequent characters). This allowed us to clean data that might be removed via deletion before the entry is complete. The second stage was once the whole entry was complete, we rolled all of the changed data back through all of the incomplete string items for this entry. This involved overlaying data replacement item information for individual strings that were wholly contained by the completed entry information, or where the replacement data fields only overlap, merging the possible replacement item information together to create a compound tag. This process was executed automatically on the study data, with no human intervention, so as to minimize the risk of leaking sensitive information. Finally, we noted that while we collected full keystroke data, only the final text data which was sent via SMS was analyzed (i.e., no partial text messages are considered). 

\section{Word Cloud Visualization}
\begin{figure}[!htb]
    \centering
 \includegraphics[width=1\columnwidth]{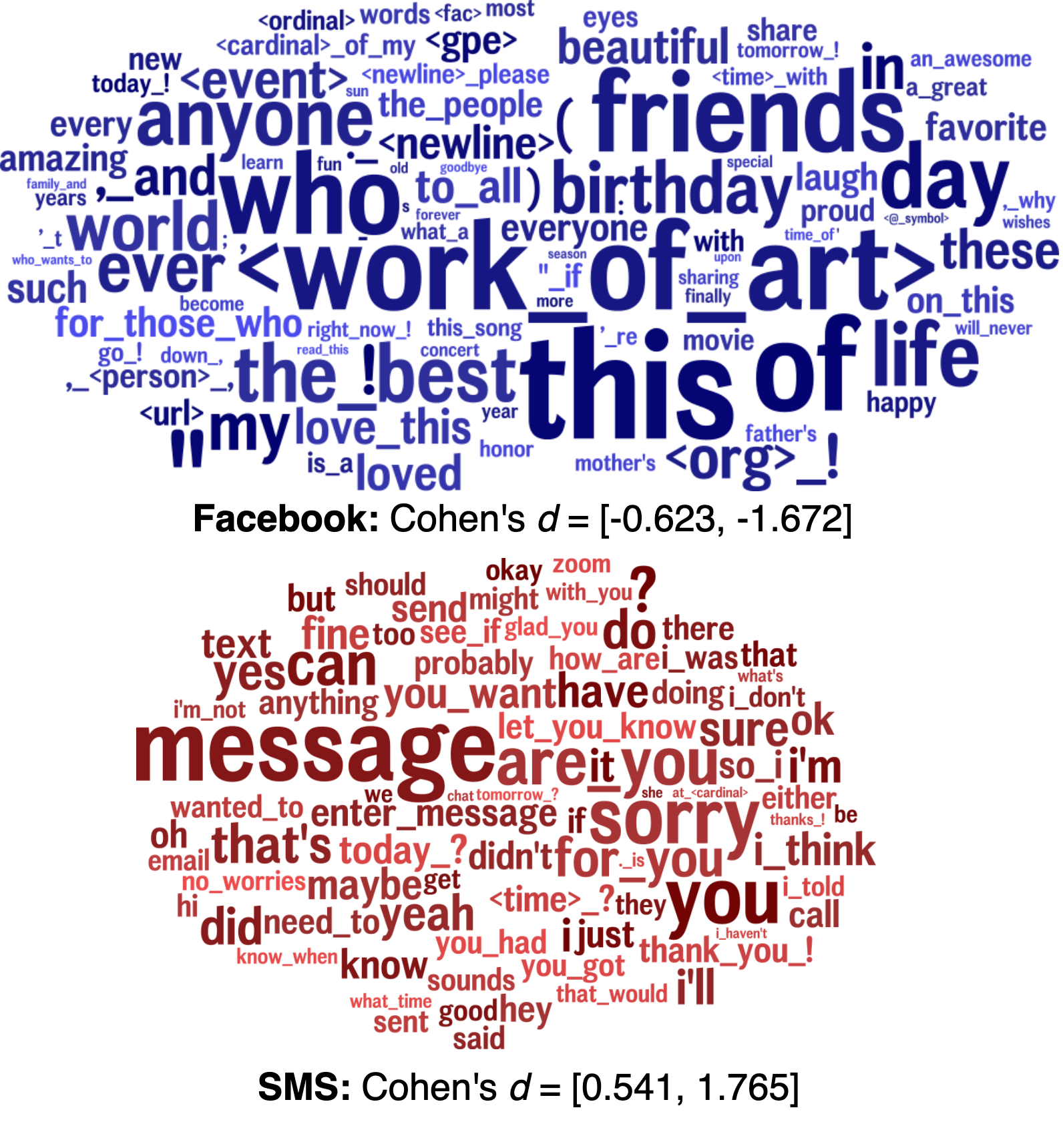}
    \caption{1-to-3 grams most correlated with Facebook vs. SMS, statistically significant at $p < 0.05$ after Benjamini-Hochberg FDR correction. Cohen's $d$ = effect size measuring Facebook vs. SMS differences. N-grams size: larger more distinguishing; darkness: darker more frequent. Angle brackets: spaCy annotated named entities (e.g., $<$work of art$>$: titles of books, songs, etc).}
    \label{fig:ngram cloud}
\end{figure}

Figure \ref{fig:ngram cloud} shows the 1-to-3 grams associated with each platform, visualized as a word cloud. 

\end{document}